\documentclass[journal]{IEEEtran}
\usepackage{ifpdf}

\usepackage{cite}
\usepackage{multirow}
\usepackage[normalem]{ulem}
\useunder{\uline}{\ul}{}

\ifCLASSINFOpdf
   \usepackage[pdftex]{graphicx}
   \usepackage[pagebackref=true,breaklinks=true,colorlinks,bookmarks=false]{hyperref}

\else
\fi
\usepackage{amsmath}
\usepackage[pagebackref=true,breaklinks=true,colorlinks,bookmarks=false]{hyperref}

\usepackage{amsfonts}
\usepackage[pagebackref=true,breaklinks=true,colorlinks,bookmarks=false]{hyperref}

\usepackage{algorithmic}
\usepackage[pagebackref=true,breaklinks=true,colorlinks,bookmarks=false]{hyperref}

\usepackage{array}
\usepackage[pagebackref=true,breaklinks=true,colorlinks,bookmarks=false]{hyperref}

\usepackage{url}
\usepackage[pagebackref=true,breaklinks=true,colorlinks,bookmarks=false]{hyperref}

\begin{document}
%

\title{Light Field Saliency Detection with Deep Convolutional Networks}
%
%
%

\author{Jun~Zhang,
        Yamei~Liu,
        Shengping~Zhang,
        Ronald~Poppe,
        and~Meng~Wang
\thanks{J. Zhang, Y. Liu, and M. Wang are with the School of Computer Science and Information Engineering, 
Hefei University of Technology, Hefei, Anhui, 230601 China.}
\thanks{S. Zhang is with the School of Computer Science and Technology, Harbin Institute of Technology, Weihai, Shandong, 264209 China.}
\thanks{R. Poppe is with the Department of Information and Computing Sciences, Utrecht University, 3584 CC Utrecht Netherlands.}
\thanks{Corresponding author: Jun Zhang (e-mail: zhangjun1126@gmail.com)}}

%
%



\maketitle


\begin{abstract}
Light field imaging presents an attractive alternative to RGB imaging because of the recording of the direction of the incoming light. The detection of salient regions in a light field image benefits from the additional modeling of angular patterns. For RGB imaging, methods using CNNs have achieved excellent results on a range of tasks, including saliency detection. However, it is not trivial to use CNN-based methods for saliency detection on light field images because these methods are not specifically designed for processing light field inputs. In addition, current light field datasets are not sufficiently large to train CNNs. To overcome these issues, we present a new Lytro Illum dataset, which contains 640 light fields and their corresponding ground-truth saliency maps. Compared to current light field saliency datasets~\cite{Li17, Zhang17}, our new dataset is larger, of higher quality, contains more variation and more types of light field inputs. This makes our dataset suitable for training deeper networks and benchmarking. Furthermore, we propose a novel end-to-end CNN-based framework for light field saliency detection. 
Specifically, we propose three novel MAC (Model Angular Changes) blocks to process light field micro-lens images.
We systematically study the impact of different architecture variants and compare light field saliency with regular 2D saliency. Our extensive comparisons indicate that our novel network significantly outperforms state-of-the-art methods on the proposed dataset and has desired generalization abilities on other existing datasets.
\end{abstract}


%
\IEEEpeerreviewmaketitle

\section{Introduction}
\begin{figure}[!t]
\centering
\includegraphics[width=1\linewidth]{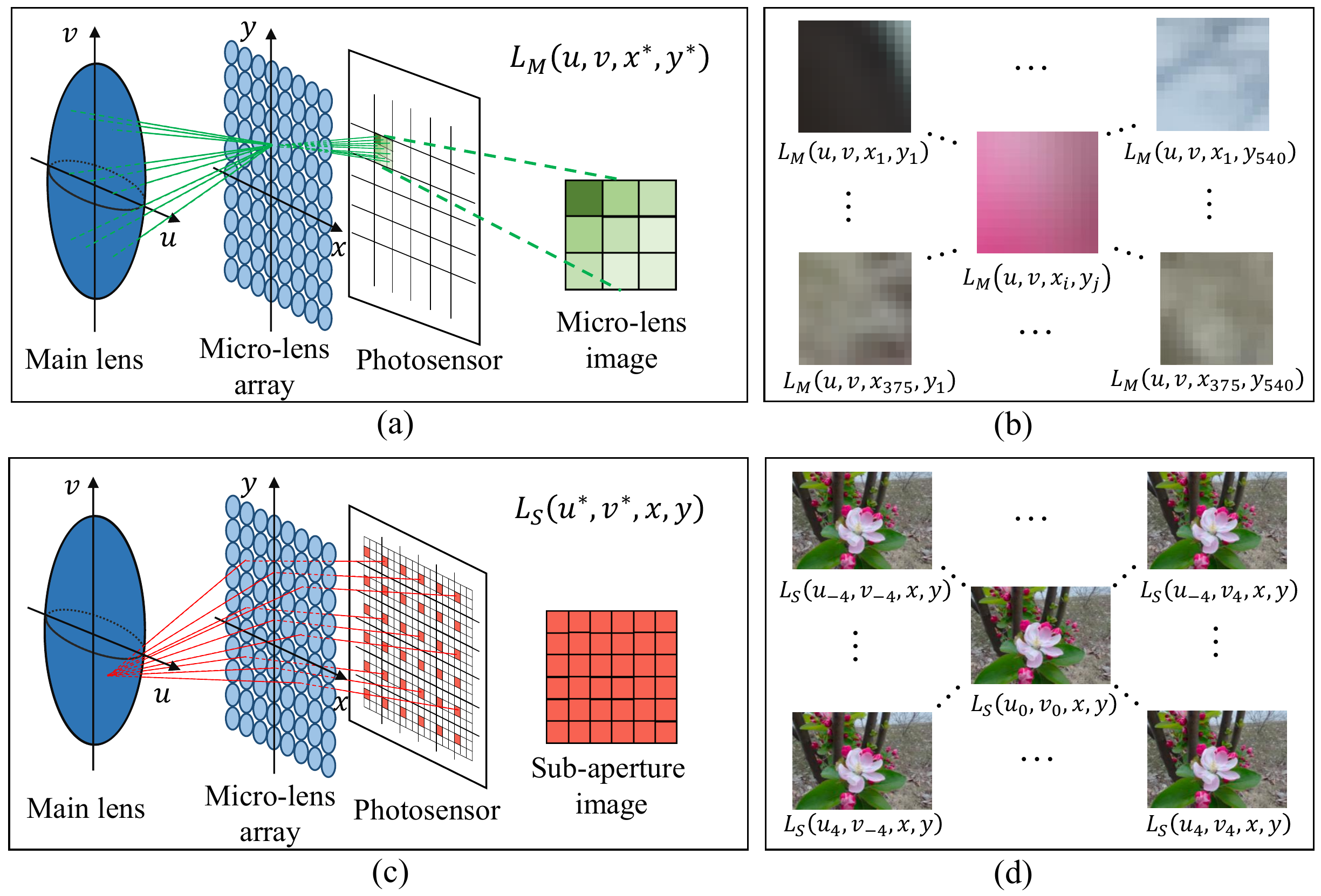}
\caption{Illustrations of light field representations. (a) Micro-lens image representation with the given location $\left (x^{*},y^{*} \right )$. (b) Micro-lens images at sampled spatial locations. (c) Sub-aperture image representation with the given viewpoint $\left (u^{*},v^{*} \right )$. (d) Sub-aperture images at sampled viewpoints, where $\left (u_{0},v_{0}\right )$ represents the central viewpoint.}
\label{fig:lf}
\end{figure}

%
%
%
%
\IEEEPARstart{L}{ight} field imaging~\cite{Adelson92} not only captures the color intensity of each pixel but also the directions of all incoming light rays.
The directional information inherent in a light field implicitly defines the geometry of the observed scene~\cite{Wanner12b}.
In recent years, commercial and industrial light field cameras with a micro-lens array inserted between the main lens and the photosensor, such as Lytro~\cite{lytro} and Raytrix~\cite{raytrix}, have taken light field imaging into a new era.
The obtained light field can be represented by 4D parameterization $\left (u,v,x,y \right )$~\cite{Levoy96}, where $uv$ denotes the viewpoint plane and $xy$ denotes the image plane, as shown in Figures~\ref{fig:lf}(a) and (c).
The 4D light field can be further converted into multiple 2D light field images, such as multi-view sub-aperture images~\cite{Levoy96}, micro-lens images~\cite{Ng05}, and epipolar plane images (EPIs)~\cite{Wanner14}.
These light field images have been exploited to improve the performance of many applications, such as material recognition~\cite{Wang16}, face recognition~\cite{Raghavendra15,Moghaddam18}, depth estimation~\cite{Tao17,Williem17,Shin18,Schilling18,Jeon18} and super-resolution~\cite{Wanner14,Yoon15,Wu19}. 

\begin{figure*}[!t]
\centering
\includegraphics[width=1\linewidth]{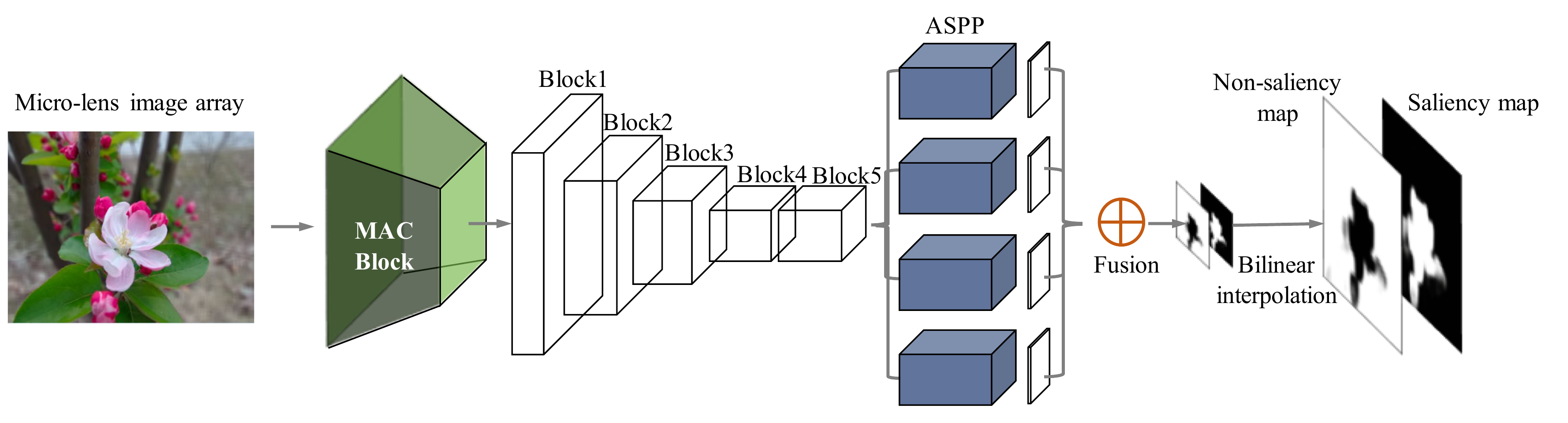}
\caption{Architecture of our network. The MAC building block converts the micro-lens image array of light fields into feature maps, which are processed by a modified DeepLab-v2 backbone model.}
\label{fig:framework}
\end{figure*}

This paper studies saliency detection on light field images. Previous work~\cite{Li17,Zhang15,Li15,Zhang17} has focused on developing hand-crafted light field features at the superpixel level by utilizing heterogenous types of light field images (\emph{e.g.,} color, depth, focusness, or flow). These methods strongly rely on low-level cues and are less capable of extracting high-level semantic concepts. This makes them unsuitable for handling highly cluttered backgrounds or predicting uniform regions inside salient objects.

In recent years, convolutional neural networks (CNNs) have been successfully applied to learn an implicit relation between pixels and salience in RGB images~\cite{Lijun15,Guanbin15,Zhao15,Liu18,Kuen16,Zhang18,Linzhao16}. These CNN-based methods have been combined with object proposals~\cite{Lijun15}, post-processing steps~\cite{Guanbin15}, contextual features~\cite{Zhao15,Liu18}, attention models~\cite{Kuen16,Zhang18}, and recurrent structures~\cite{Linzhao16}.
Although these approaches achieve improved performance over saliency detection on benchmark datasets, they often adopt complex network architectures, which limits generalization and complicates training.
Besides, the limited information in RGB images does not allow to fully exploit geometric constraints, which have been shown to be beneficial in saliency detection~\cite{Niu12,Sitzmann18}.

In this paper, we propose a novel method to predict the salience of light fields by utilizing deep learning technologies. 
Even with the emergence of CNNs for RGB images, there are still two key issues for saliency detection on light field images: 
(1) {\bf Dataset.} The RGB datasets~\cite{Achanta09,Yang13,Yin14} are not sufficient to address significant variations in illumination, scale and background clutter.
Previous light field saliency datasets LFSD~\cite{Li17} and HFUT-Lytro~\cite{Zhang17} include only 100 and 255 light fields, respectively, captured by the first-generation Lytro cameras. They are not large enough to train deep convolutional networks without severely overfitting. In addition, the unavailability of multi-views in the LFSD dataset and the color distortion of the sub-aperture images in the HFUT-Lytro dataset impede an evaluation of existing methods.
(2) {\bf Architecture.} The adoption of CNN-based architectures in light field saliency detection is not trivial because the existing CNNs for 2D images do not support the representation of 4D light field data.
Thus, novel architectures must be developed for saliency detection in light field images.

Based on the aforementioned issues, we introduce a comprehensive, realistic and challenging benchmark dataset for light field saliency detection.
Using a Lytro Illum camera, we collect $640$ light fields with significant variations in terms of size, amount of texture, background clutter and illumination.
For each light field, we provide a high-quality micro-lens image array that contains multiple viewpoints for each spatial location. 

To explore spatial and multi-view properties of light fields for saliency detection, we further propose a novel deep convolutional network based on the modified DeepLab-v2 model~\cite{Chen18a} as well as several architecture variants (termed as {\it MAC} blocks) specifically designed for light field images.
Our blocks aim to Model Angular Changes (MAC) from micro-lens images in an explicit way.
One block type is similar to the angular filter explored in ~\cite{Wang16} for material recognition.
The main difference is that ~\cite{Wang16} reorganizes the 4D information of the light field in different ways and divide the input into patches, which are further processed by the VGG-16 network for patch classification. 
We observe that this study does not pay much attention to the angular changes that may arise due to the network parameters, as it lacks an in-depth analysis of the relationship between its learned features and reflected information beneficial to material recognition.
In contrast, inspired by the micro-lens array hardware configuration of light field cameras, the proposed MAC blocks are specially tailored to process micro-lens images in an explicit way.
The network parameters are designed to sample different views and capture view dependencies by performing non-overlapping convolution on each micro-lens image.
We experimentally show that the angular changes are consistent with the viewpoint variations of micro-lens images, and the effective angular changes of each pixel may increase depth selectivity and the ability for accurate saliency detection.
Figure~\ref{fig:framework} provides an overview of the proposed network.

Our contributions are summarized as follows:
\begin{itemize}
\item We construct a new light field dataset for saliency detection, which comprises of $640$ high-quality light fields and the corresponding per-pixel ground-truth saliency maps.
This is the largest light field dataset that enables efficient deep network training for saliency detection, and addresses new challenges in saliency detection such as inconsistent illumination and small salient objects in the cluttered or similar background.
\item We propose an end-to-end deep convolutional network for predicting saliency on light field images. To the best of our knowledge, no work has been reported on employing deep learning techniques for light field saliency detection.
\item We provide an analysis of the proposed architecture variants specifically designed for light-field inputs. We also quantitatively and qualitatively compare our best-performing architecture with the 2D model using the central viewing image and other 2D RGB-based methods. We show that our network outperforms state-of-the-art methods on the proposed dataset and generalizes well to other datasets.
\end{itemize}

The remainder of this paper is structured as follows. The next section summarizes related work on light field datasets, saliency detection from light field images, and saliency detection using deep learning technologies. We introduce our novel Lytro Illum saliency dataset in Section~\ref{sec:dataset}. We introduce our novel MAC blocks in Section~\ref{network} and evaluate them in Section~\ref{sec:evaluation}. We conclude in Section~\ref{sec:conclusion}.

\section{Related work}
\subsection{Light field datasets for saliency detection}
There are only two existing datasets designed for light field saliency detection, both recorded with Lytro's first-generation cameras. The Light Field Saliency Database (LFSD)~\cite{Li17} contains $100$ light fields with $360\times 360$ spatial resolution. A rough focal stack and an all-focus image are provided for each light field.
The images in this dataset usually have one salient foreground object and a background with good color contrast.
The limited complexity of the dataset is not sufficient to address the variety of challenges for saliency detection when using a light field camera, such as illumination variations and small objects on the similar or cluttered background.
Later, Zhang \emph{et al.}~\cite{Zhang17} proposed the HFUT-Lytro dataset, which consists of $255$ light fields with complex backgrounds and multiple salient objects.
Each light field has a $7\times 7$ angular resolution and $328\times 328$ pixels of spatial resolution.
Focal stacks, sub-aperture images, all-focus images, and coarse depth maps are provided in this dataset.
However, the color channels in their sub-aperture images are distorted owing to the under-sampling during decoding~\cite{Dansereau13}.
In this work, we use a Lytro Illum camera to build a larger, higher-quality and more challenging saliency dataset by capturing more variations in illuminance, scale, position.
We also generate the micro-lens image array for each light field, which is not provided in previous datasets.

\subsection{Saliency detection on light field images}
Previous methods for light field saliency detection rely on superpixel-level hand-crafted features~\cite{Li14,Li17,Li15,Zhang15,Zhang17}.
Pioneering work by Li \emph{et al.}~\cite{Li14,Li17} shows the feasibility of detecting salient regions using all-focus images and focal stacks from light fields.
Zhang \emph{et al.}~\cite{Zhang15} explored the light field depth cue in saliency detection, and further computed light field flow fields over focal slices and multi-view sub-aperture images to capture depth contrast~\cite{Zhang17}.
In~\cite{Li15}, a dictionary learning-based method is presented to combine various light field features using a sparse coding framework. 
Notably, these approaches share the assumption that dissimilarities between image regions imply salient cues.
In addition, some of them~\cite{Li15,Zhang15,Zhang17} also utilize refinement strategies to enforce neighboring constraints for saliency optimization.
In contrast to the above methods, we propose a deep convolutional network by learning efficient angular kernels without additional refinement on the upsampled image.

\subsection{Deep learning for saliency prediction}
Recently, remarkable advances in deep learning drive research towards the use of CNNs for saliency detection~\cite{Lijun15,Guanbin15,Zhao15,Kuen16,Zhang18,Liu18,Linzhao16}.
Different from conventional learning-based methods, CNNs can directly learn a mapping between 2D images and saliency maps. 
Since the task is closely related to pixel-wise image classification, most works have built upon successful architectures for image recognition on the ImageNet dataset~\cite{Russakovsky15}, often initializing their networks with the VGG network~\cite{Simonyan14}.
For example, several methods directly use CNNs to learn effective contextual features and combine them to infer saliency~\cite{Zhao15,Liu18}.
Other methods extract features at multiple scales and generate saliency maps in a fully convolutional way~\cite{Li16,Kruthiventi16}. 
Recently, attention models~\cite{Kuen16,Zhang18} have been introduced to saliency detection to mimic the visual attention mechanism by focusing on informative regions in visual scenes. 
Another direction for improving the quality of the saliency maps is the use of a recurrent structure~\cite{Linzhao16}, which mainly serves as a refinement stage to correct previous errors.
Although deep CNNs have achieved great success in saliency detection, none of them addresses challenges in the 4D light field.
Directly applying the existing network architectures to light field images would not be appropriate because a standard network is not particularly good at capturing viewpoint changes in light fields.
Our work is the first to address light field saliency detection with end-to-end deep convolutional networks.

\subsection{Deep learning technologies on light field data}
Recently, in terms of different light field image types, learning-based techniques have been explored for light field image processing.
Yoon \emph{et al.}~\cite{Yoon15} proposed a deep learning framework for spatial and angular super-resolution, in which two adjacent sub-aperture images are employed to generate the in-between view. 
Wang \emph{et al.}~\cite{Wang18b} built a bidirectional recurrent CNN to super-resolve horizontally and vertically adjacent sub-aperture image stacks separately and then combined them using a multi-scale fusion scheme to obtain complete view images.
Very recently, Zhang \emph{et al.}~\cite{Zhang19} designed a residual network structure to process one central view image and four stacks of sub-aperture images from four angular directions.
Residual information from different directions is then combined to yield the high-resolution central view image.
Kalantari \emph{et al.}~\cite{Kalantari16} proposed the first deep learning framework for view synthesis.
They applied two sequential CNNs on only four corner sub-aperture images to model depth and color estimation simultaneously by minimizing the error between synthesized views and ground truth images.
Wu \emph{et al.}~\cite{Wu19} introduced the “blur-restoration-deblur” framework for light field reconstruction on 2D EPIs (epipolar plane images).
In order to directly synthesize novel views of dense 4D light fields from sparse views, Wang \emph{et al.}~\cite{Wang18} assembled 2D strided convolutions operated on stacked EPIs and two detail-restoration 3D CNNs connected with angular conversion to build a pseudo 4D CNN.
Heber \emph{et al.}~\cite{Heber16} applied a CNN in a sliding window fashion for shape from EPIs, which allows to estimate the depth map of a predefined sub-aperture image.
In the successive work~\cite{Heber17}, they designed a U-shaped network for disparity estimation operating on a EPI volume with two spatial dimensions and one angular dimension.
The most similar to our work is ~\cite{Wang16}, which proposes several CNN architectures based on different light field image types.
One of the architectures is developed for the images similar to raw micro-lens images.
However, the network is mainly designed to verify the advantages of multi-view information of light field compared with 2D RGB images in material recognition.
In contrast, our work specifically focuses on the learned angular features from micro-lens images and their relationship with salient/non-salient cues.

\section{The Lytro Illum saliency dataset} \label{sec:dataset}
To train and evaluate our network for saliency detection, we introduce a comprehensive novel light field dataset.
\begin{figure*}[!t]
\centering
\includegraphics[width=1\linewidth]{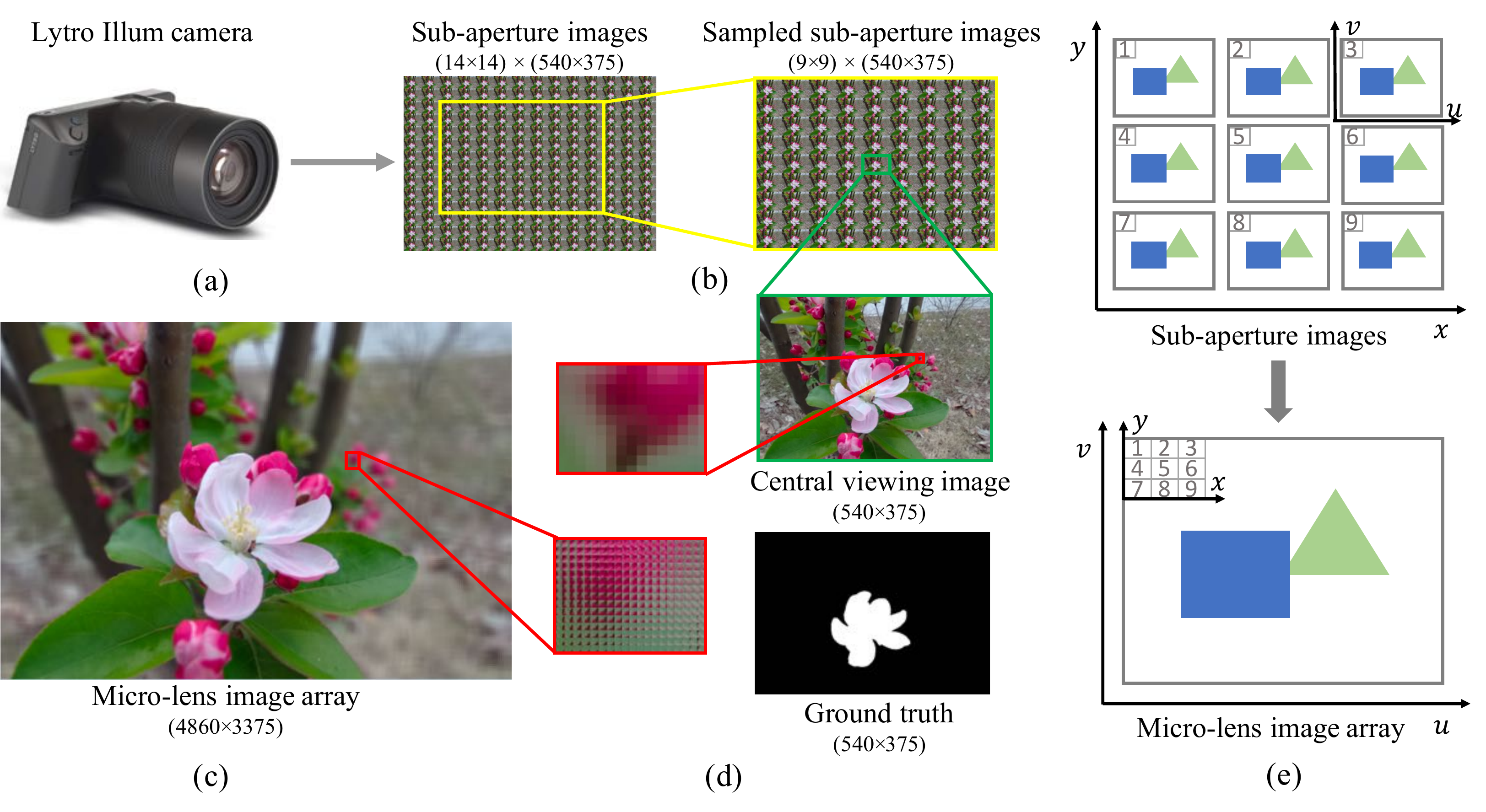}
\caption{Flowchart of the dataset construction. (a) Lytro Illum camera. (b) Sub-aperture images. (c) Micro-lens image array. (d) Ground-truth map for the central viewing image. (e) The generation of a micro-lens image array from sub-aperture images. The digits indicate viewpoints.}
\label{fig:dataset}
\end{figure*}

\subsection{Light field representation}
\label{representation}
There are various ways to represent the light field~\cite{McMillan95b,Adelson91,Levoy96}. We adopt the two-plane parameterization~\cite{Levoy96} to define the light field as a 4D function $L\left (u,v,x,y \right )$, where $u\times v$ indicates the angular resolution and $x\times y$ indicates the spatial resolution.
As illustrated in Figures~\ref{fig:lf}(a) and (b), a set of all incoming rays from the $uv$ plane intersected with a given micro-lens location $\left (x^{*},y^{*} \right )$ produces a micro-lens image with multiple viewpoints $L_{M}\left ( u,v,x^{*},y^{*} \right )$.
The micro-lens images from different locations can be arranged into a micro-lens image array.
As shown in Figures~\ref{fig:lf}(c) and (d), all micro-lens regions on the $xy$ plane receive the incoming rays from a given angular position $\left (u^{*},v^{*} \right )$, which produces a sub-aperture image with all locations $L_{S}\left (u^{*},v^{*},x,y \right )$.
The central viewing image is formed by the rays passed through the main lens optical center $\left (u=u_{0}, v=v_{0}\right )$.
Since the sub-aperture images contain optical distortions caused by the light rays passed through the lens~\cite{Tang13,Jeon15}, in this paper, we build our network based on the micro-lens images, which have been shown advantages over the sub-aperture images for scene reconstruction~\cite{Shuo18}.

\subsection{Dataset construction}
\label{construction}
Figure~\ref{fig:dataset} illustrates the procedure of our light field dataset construction.
First, a set of 4D light fields are obtained using a Lytro Illum camera (Figure~\ref{fig:dataset}(a)). Second, we use Lytro Power Tools (LPT) \cite{lpt} to decode light fields from raw 4D data to 2D sub-aperture images so that each light field has a spatial resolution of $540\times 375$ and an angular resolution of $14\times 14$.
To reach a compromise on the training time and the detection accuracy, we sample $9\times 9$ viewpoints from each light field to generate new sub-aperture images, as shown in Figure~\ref{fig:dataset}(b).
Third, we generate a micro-lens image by sampling the same spatial location from each sub-aperture image (see Figure~\ref{fig:dataset}(e)), which further produces a micro-lens image array of size $4860\times 3375$, shown in Figure~\ref{fig:dataset}(c).
The red region indicates one pixel with $9\times 9$ observation viewpoints in Figure~\ref{fig:dataset}(c), comparing to one pixel only with the central view in Figure~\ref{fig:dataset}(d).
We initially collect 800 light fields and manually annotate the per-pixel ground-truth label for each central viewing image.
To reduce label inconsistency, each image is annotated by five independent annotators.
We only regard a pixel as salient if it is verified by at least three annotators. We only keep those images with sufficient agreement. In the end, our new dataset contains $640$ light fields with $81$ views. 

\begin{figure*}[!t]
\centering
\includegraphics[width=1\linewidth]{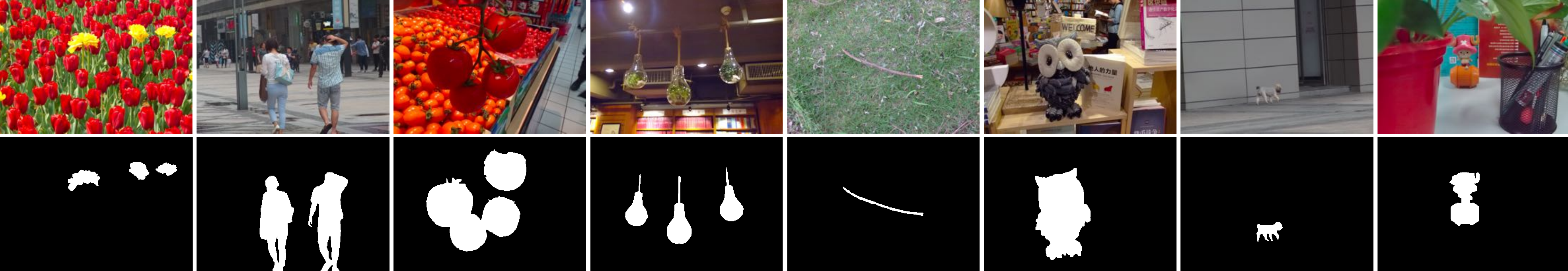}
\caption{Example central viewing images (top) and their corresponding ground-truth saliency maps (bottom) from our novel Lytro Illum dataset.}
\label{fig:examples}
\end{figure*}

Figure~\ref{fig:examples} shows eight examples of central viewing images and their corresponding ground-truth saliency maps.
There are significant variations in illumination, spatial distribution, scale and background.
Besides, there are multiple regions for some saliency annotations.

\begin{figure*}[!t]
\centering
\includegraphics[width=1\linewidth]{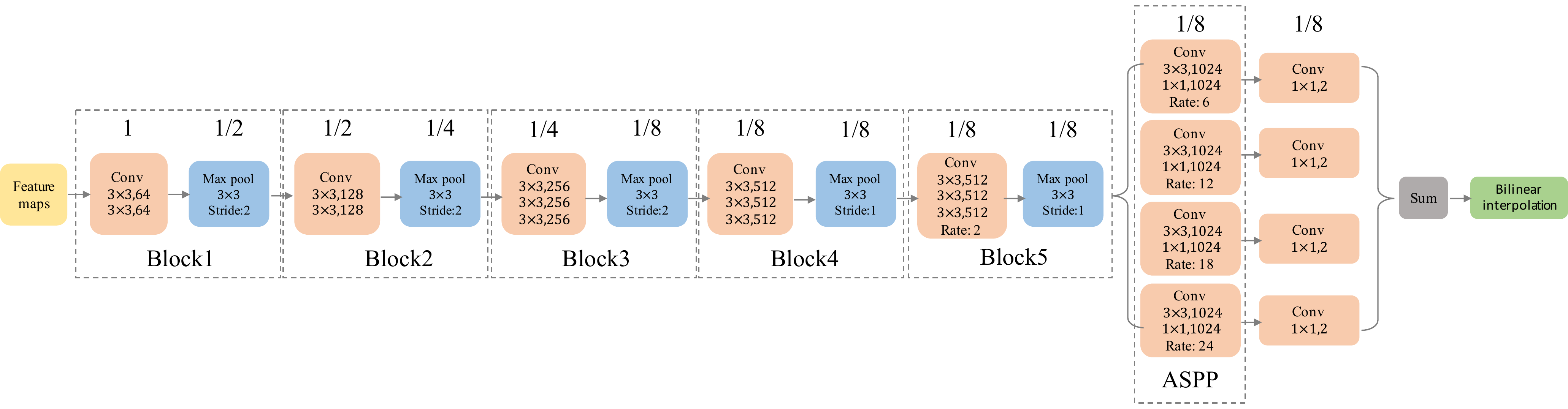}
\caption{Network structure of the backbone model based on DeepLab-v2~\cite{Chen18a}. The reduction in resolution is shown at the top of each box.}
\label{fig:deeplab}
\end{figure*}

\section{Light field saliency network}
\label{network}
We propose an end-to-end deep convolutional network framework for light field saliency detection as shown in Figure~\ref{fig:framework}.
Based on the micro-lens image array, the MAC (Modal Angular Changes) blocks are designed to transfer the light field inputs to feature maps in different ways.
Then, the feature maps are fed to a modified DeepLab-v2~\cite{Chen18a} to predict saliency maps. We first discuss the backbone model and then detail different MAC block variants.

\subsection{Backbone model}
\label{deeplab}
We formulate light field saliency detection as a binary pixel labeling problem.
Saliency detection and semantic segmentation are closely related because both are pixel-wise labeling tasks and require low-level cues as well as high-level semantic information.
Inspired by previous literature on semantic segmentation~\cite{Shelhamer17,Chen16,Chen18a}, we design our backbone model based on DeepLab~\cite{Chen16}, which is a variant of FCNs~\cite{Shelhamer17} modified from the VGG-16 network~\cite{Simonyan14}.
There are several variants of DeepLab~\cite{Chen18a,Szegedy16,Chen18}. 
In this work, we use DeepLab-v2~\cite{Chen18a}, which introduces ASPP to capture multi-scale information and long-range spatial dependencies among image units.

The modified network is composed of five convolutional ({\it conv}) blocks, each of which is divided into convolutions followed by a ReLu.
A max-pooling layer is connected after the top {\it conv} layer of each {\it conv} block.
The ASPP is applied on top of block5, which consists of four branches with atrous rates ($r=\left \{6,12,18,24 \right \}$).
Each branch contains one $3 \times 3$ convolution and one $1 \times 1$ convolution.
The resulting features from all branches are then passed through another $1 \times 1$ convolution and summed to generate the final score.
The network further employs bilinear interpolation to upsample the fused score map to the original resolution of the central viewing image, which produces the saliency prediction at the pixel level.
In addition, we add dropout to all the {\it conv} layers of the five blocks to avoid overfitting and set the $1 \times 1$ {\it conv} layer with $2$ channels after ASPP to produce saliency and non-saliency score maps.
The detailed architecture is illustrated in Figure~\ref{fig:deeplab}.

\begin{figure*}[!t]
\centering
\includegraphics[width=1\linewidth]{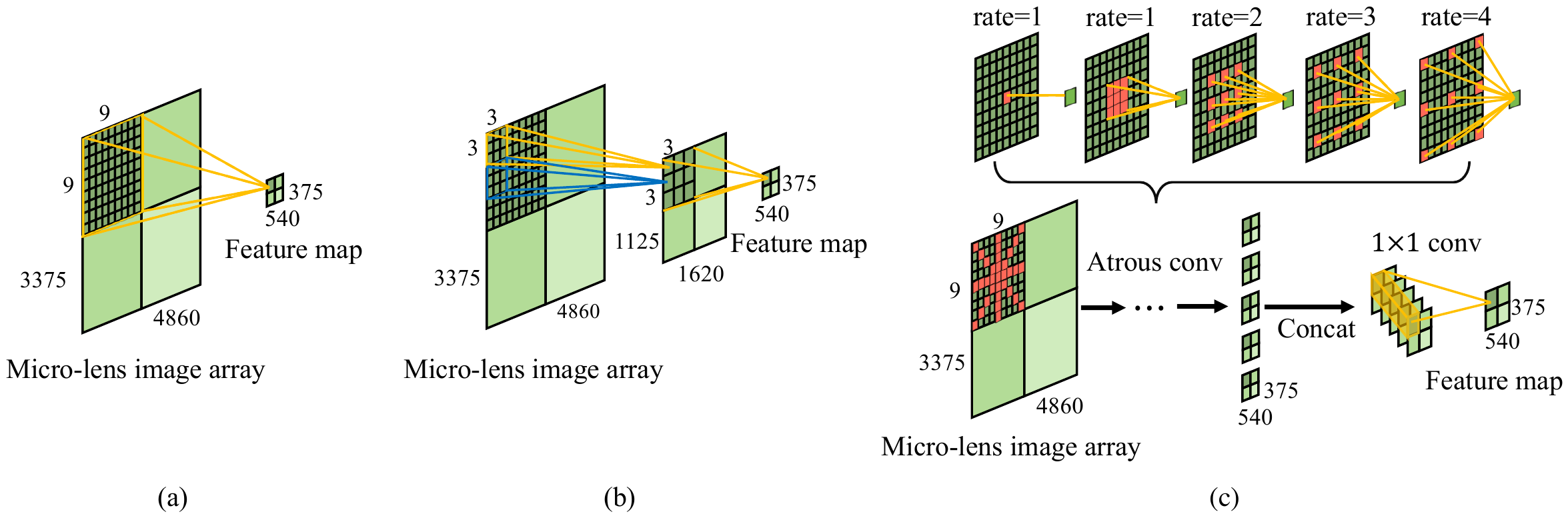}
\caption{Architectures of the proposed MAC blocks. (a) MAC block-{\bf $9 \times 9$}. (b) MAC block-{\bf $3 \times 3$}. (c) MAC block-star shaped. The selected viewpoints are highlighted in red.}
\label{fig:variants}
\end{figure*}

\subsection{MAC blocks}
\label{variants}
Our network is essentially a modified DeepLab-v2 network augmented with a light field input process.
As shown in Figure~\ref{fig:framework}, a MAC block is a basic computational unit operating on a micro-lens image array input $\mathbf{M}\in \mathbb{R}^{W\times H\times C}$ and producing an output feature map $\mathbf{F}\in \mathbb{R}^{{W}'\times {H}'\times {C}'}$.
Here, $W=N_x\times N_u$ and $H=N_y\times N_v$, in which $(N_x, N_y)$ is the spatial size and $(N_u, N_v)$ is the view size, respectively.
The motivation of the MAC block is to model angular changes at one pixel location in an explicit manner.
An essential part of learning angular features is the design of convolutional kernels applied on the micro-lens images. 
However, it is unclear what defines “good” angular filters and how many angular directions should be chosen for better performance.
In this paper, we propose three different MAC block variant architectures to process light field micro-lens image arrays before block1 of the backbone model, in which convolutional methods with kernel sizes, stride size and sampled viewpoints are all designed to capture angular changes in light fields, as shown in Figure~\ref{fig:variants}.

For the design simplicity of the MAC block, some default settings are fixed to guarantee that the predicted map and the ground truth map have the same spatial resolution in a fully convolutional network architecture.
First, the spatial dimension of output of the MAC block is ensured to be the same with that of the 2D sub-aperture image, \emph{i.e.} ${W}'=N_x$ and ${H}'=N_y$.
Second, the number ${C}'$ of convolutional kernels in the MAC block is the same as that of convolution kernels in block1 of DeepLab-v2.
In our case where the data are captured by a Lytro Illum camera, the MAC block converts the light field input data into a $540\times 375\times 64$ feature map.
The parameters of LFNet variants, including the kernel size $k\times k\times C$ and the convolutional stride $s$, should meet the above two conditions.
We now discuss the detailed architectures of the three proposed MAC blocks.

\subsubsection{MAC block-{\bf $9\times 9$}}
As described in Section~\ref{construction},  each micro-lens image has $9\times 9$ viewpoints and can be considered as one of the pixel locations.
The spatial resolution is $540\times 375$ thus the size of the whole micro-lens image array is $4860\times 3375$.
In this architecture, we design angular convolutional kernels across all viewpoint directions, as shown in Figure~\ref{fig:variants}(a).
The kernel size shares the same angular resolution of one micro-lens image, and the number of kernels and the stride size are set to extract angular features for each micro-lens image.
Specifically, we propose 64 angular kernels, each of which is a $9 \times 9$ filter.
The stride of convolution operations is $9$, which leads to $540 \times 375 \times 64$ feature maps.
Each point on the feature map can be considered as being captured by the $81$ lenslets.
These kernels differ from common convolutional kernels applied on 2D images in that they only detect the angular changes in the micro-lens image array.
This architecture directly learns the angular information from light field images, and thus is expected to distinguish salient foregrounds and backgrounds with similar colors or textures.

\subsubsection{MAC block-{\bf $3\times 3$}}
Motivated by the effectiveness of the smaller kernels in VGG-16~\cite{Simonyan14} and Inception v2~\cite{Szegedy16}, 
we replace the $9\times 9$ convolution in MAC block-$9\times 9$ with two layers of $3\times 3$ convolution (stride$=3$) shown in Figure~\ref{fig:variants}(b), which increases the number of parameters while enhancing the network nonlinearity.

\subsubsection{MAC block-star shaped}
We design atrous angular convolutional kernels to capture long-range angular features. 
The atrous rates are set to sample representative viewpoint directions. It has been shown that using selected angular directions is beneficial in the context of depth estimation~\cite{Strecke17,Shin18}. Here, we test the application in saliency detection. Different from MAC block-$9 \times 9$, we select star-shaped viewpoints (\emph{i.e.} four directions $\theta= \left \{0^{\circ},45^{\circ},90^{\circ},135^{\circ} \right \}$) from each micro-lens image.
To implement viewpoint sampling and angular filtering, we use atrous convolution with five atrous rates, as shown in Figure~\ref{fig:variants}(c).
The resulting feature maps are concatenated and combined using $1\times 1$ convolutions for later processing.

{\bf Adaptation.}
Note that although the proposed framework is specially tailored to process micro-lens based light field data for saliency detection, in theory, the proposed network could be adapted to different types of light field data as well if the camera acquires the same pixel with sufficient different views.
However, directly using the proposed network to process other types of light field data potentially has the following problems. 
(i) The number of views is usually limited and the resulting images suffer from angular aliasing due to the poor angular sampling of other cameras, such as multi-camera arrays~\cite{Wilburn05} and light field gantries~\cite{Wanner13}. 
(ii) For sparse and wide-baseline light fields captured by multi-camera arrays, convolution operating over the full resolution of light fields may be prohibitively memory intensive and computationally expensive. 
(iii) Compared to micro-lens based light field cameras offered by simple design, flexibility and little marginal costs, other light field systems are generally impractical for the outdoor data collection due to the complex, heavy, and cost of the capturing system. 
It is difficult to construct a realistic and challenging saliency dataset of a certain scale. 
Therefore, the proposed network in this paper is currently only applicable to micro-lens based light field data with a set of dense views.

\section{Experimental Results} \label{sec:evaluation}
Our experimental evaluation is split up into three main parts.
The first section evaluates the three variations of the MAC blocks to identify the network design that works best for light-field saliency detection.
The second section discusses the angular resolution, the overfitting issue and the advantages of the selected variant of the MAC block compared to 2D saliency detectors.
Finally, the third section shows the performance we can attain based on the best performing model.
We show state-of-the-art results on the new Lytro Illum, HFUT-Lytro~\cite{Zhang17} and LFSD datasets~\cite{Li17}, based on a pre-trained network on the proposed dataset.

\subsection{Settings}
\subsubsection{Implementation and training}
The computational environment has an Intel i7-6700K CPU@4.00GHz, 15GB RAM, and an NVIDIA GTX1080Ti GPU.
We trained our network using the Caffe library~\cite{jia2014caffe} with the maximum iteration step of 160K. 
We initialize the backbone model with DeepLab-v2~\cite{Chen18a} pre-trained on the PASCAL VOC 2012 segmentation benchmark~\cite{Everingham14}.
The newly added {\it conv} layers in the MAC block, the first layer of block1, and the score layer are initialized using the Xavier algorithm~\cite{Glorot10}.
The whole network is trained end-to-end using the stochastic gradient descent (SGD) algorithm.
To leverage the training time and the image size, we use a single image batch size.
Momentum and weight decay are set to $0.9$ and $0.0005$, respectively.
The base learning rate is initialized as $0.01$ for the newly added {\it conv} layers in the MAC block and the first layer of block1, and $0.001$ with the poly decay policy for the remaining layers.
A dropout layer with probabilities $p=\left [ 0.1,0.1,0.2,0.2,0.3,0.5 \right ]$ is applied after {\it conv} layers for block1--block5 and ASPP, respectively.

We use the softmax loss function defined as
\begin{equation}
\begin{aligned}
L= -\frac{1}{W\times H}\sum_{i=1}^{W}\sum_{j=1}^{H}log\frac{e^{z_{i,j}^{y_{i,j}}}}{e^{z_{i,j}^{0}}+e^{z_{i,j}^{1}}}
\end{aligned}
\end{equation}
where $W$ and $H$ indicate the width and height of an image, $z_{ij}^{0}$ and $z_{ij}^{1}$ are the last two activation values of the pixel $(i,j)$ and $y_{ij}$ is the ground-truth label of the pixel $(i,j)$.
Note that $y_{ij}$ is 1 only when pixel $(i,j)$ is salient.
Our code and dataset are available at \url{https://github.com/pencilzhang/MAC-light-field-saliency-net.git}.

\subsubsection{Datasets}
\begin{figure}[!t]
\centering
\includegraphics[width=1\linewidth]{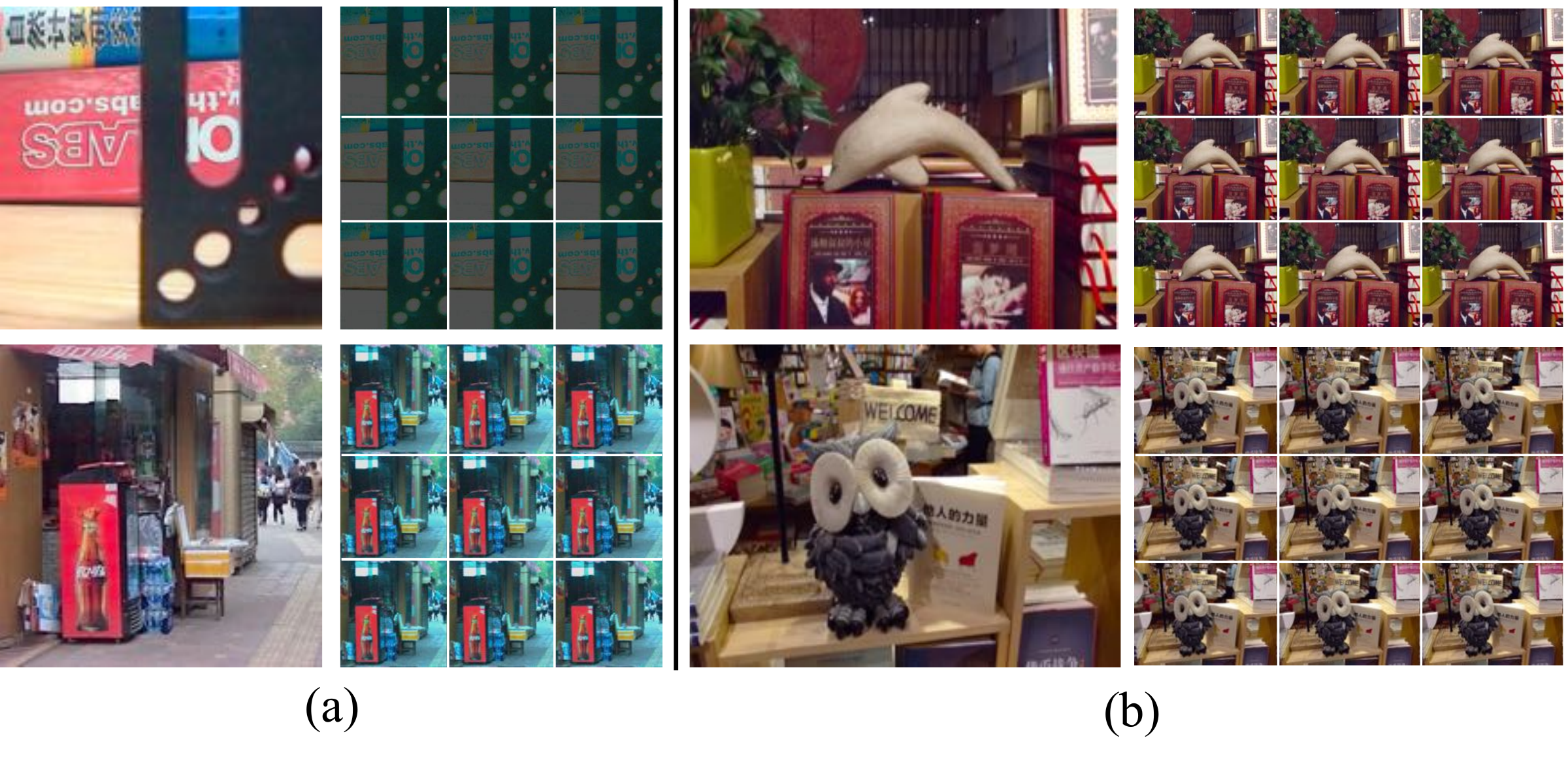}
\caption{Light field image examples.  (a) The HFUT-Lytro dataset. (b) The proposed Lytro Illum dataset. Left: the all-focus images and the central viewing images are shown for the two datasets, respectively. Right: nine sub-aperture images are randomly sampled for each light field.}
\label{fig:lytro_data}
\end{figure}

Three datasets are used for benchmarking: the proposed Lytro Illum dataset, the HFUT-Lytro dataset~\cite{Zhang17}, and the LFSD dataset~\cite{Li17}.
Our network is trained and evaluated on the proposed Lytro Illum dataset using a five-fold cross-validation.
The trained model is further tested on the other two datasets to evaluate the generalization ability of our network.
Note that the unavailable viewpoints in the LFSD dataset and the color distortion of sub-aperture images in the HFUT-Lytro dataset (see examples in Figure~\ref{fig:lytro_data} for visual comparison) are unsuitable for evaluation of our method.
To apply the trained model on the two datasets, we pad the angular resolutions to $9\times 9$ using the all-focus image.

\subsubsection{Data augmentation}
In order to obtain more training data to achieve good performance without overfitting,
we augment the training data aggressively on-the-fly.
To facilitate this augmentation, we use geometric transformations (\emph{i.e.} rotation, flipping and cropping), changes in brightness, contrast, and chroma as well as additive Gaussian noise.
Specifically, we rotate the micro-lens image array $90$, $180$, and $270$ degrees, and perform horizontal and vertical flipping.
To change the relative position of the saliency region in the image, we randomly crop two subimages of $3519\times 2907$ size from the micro-lens image array.
Then for one subimage and the image arrays with $0$, $90$, and $180$ degrees of rotation, we adjust the brightness by multiplying all pixels by $1.5$ and $0.6$, respectively, and both chroma and contrast by the multiplication factor $1.7$.
Finally, we add the zero-mean Gaussian noise with variance of $0.01$ to all images.
In total, we expand the micro-lens image array by 48 ($\left (4\times 4+8 \right )\times 2$) such that the whole training dataset is increased from $512$ to $24,576$.

\subsubsection{Evaluation metrics}
We adopt five metrics to evaluate our network. 
The first one is precision-recall (PR) curve. 
Specifically, saliency maps are first binarized and then compared to the ground truths under varying thresholds.
The second metric is $F_{\beta}$--measure, which considers both precision and recall
\begin{equation}
F_{\beta}=\frac{(1+\beta ^{2})Precision\cdot Recall}{\beta ^{2}\cdot Precision+ Recall}
\end{equation}
where $\beta^2$ is set to $0.3$ as suggested in~\cite{Achanta09}.
The third metric is Average Precision (AP), which is computed by averaging the precision values at evenly spaced recall levels.
The fourth metric is Mean Absolute Error (MAE), which computes the average absolute per-pixel difference between the predicted map and the corresponding ground truth map.
Additionally, to amend several limitations of the above four metrics, such as interpolation flaw for AP, dependency flaw for PR curve and $F_{\beta}$--measure, and equal-importance flaw for all metrics, as suggested in~\cite{Margolin14}, we use weighted $F_{\beta }^{w}$ (WF)--measure based on weighted precision and recall as the fifth metric
\begin{equation}
F_{\beta}^{w}=\frac{(1+\beta ^{2})Precision^{w}\cdot Recall^{w}}{\beta ^{2}\cdot Precision^{w}+ Recall^{w}}
\end{equation}
where $w$ is a weighting function based on the Euclidean distance to calculate the pixel importance from the ground truth.

\subsection{Evaluation of MAC blocks}
\label{ablation}
We present a detailed performance comparison among different MAC block variant architectures on the proposed Lytro Illum dataset.
As described in Section~\ref{variants}, these variants only differ in the convolution operations applied on their light field inputs.
The quantitative results of the comparison are shown in Table~\ref{MAC}, from which we can see that the MAC block-$9\times 9$ architecture achieves the best performance for all metrics on the proposed dataset.
We hypothesize that treating every micro-lens image as a whole and applying the angular kernels that have the same size with the angular resolution of the light field can help to exploit the multi-view information in the micro-lens image array.
The detection performances of two other variants are lower, probably because the increased number of parameters make the network more difficult to train.
\begin{table}[!t]
\renewcommand{\arraystretch}{1.3}
\caption{Quantitative results on the proposed Lytro Illum dataset}
\label{MAC}
\centering
\begin{tabular}{|l|c|c|c|c|c|}
\hline
Method                                         & F-measure       & WF-measure      & MAE             & AP              \\
\hline
MAC block--star shaped                             & 0.8045          & 0.7426          & 0.0555    & 0.9120          \\
MAC block--$3\times 3$                   & 0.8066          & 0.7471          & 0.0562          & 0.9118          \\
MAC block--$9\times 9$                  & \textbf{0.8116}    & \textbf{0.7540}    & \textbf{0.0551} & \textbf{0.9124}    \\
\hline
\end{tabular}
\end{table}

\begin{figure}[!t]
\centering
\includegraphics[width=1\linewidth]{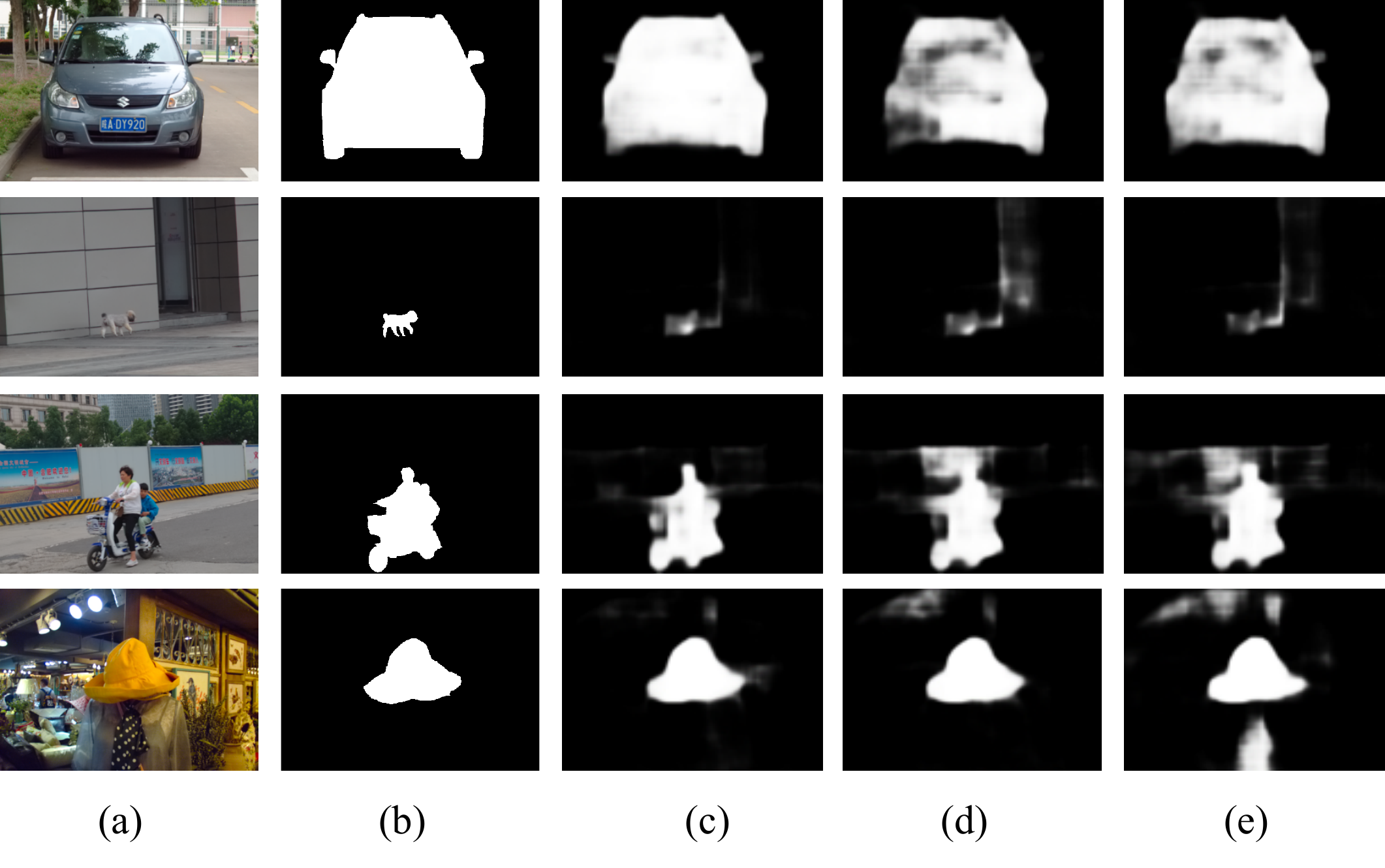}
\caption{Visual comparison of different MAC block variants. (a) Central viewing images. (b) Ground truth maps. (c) MAC block-$9\times 9$. (d) MAC block-$3\times 3$. (e) MAC block-star shaped.}
\label{fig:vis_ours}
\end{figure}

Figure~\ref{fig:vis_ours} presents qualitative results of all variants.
\begin{figure*}[!t]
\centering
\includegraphics[width=1\linewidth]{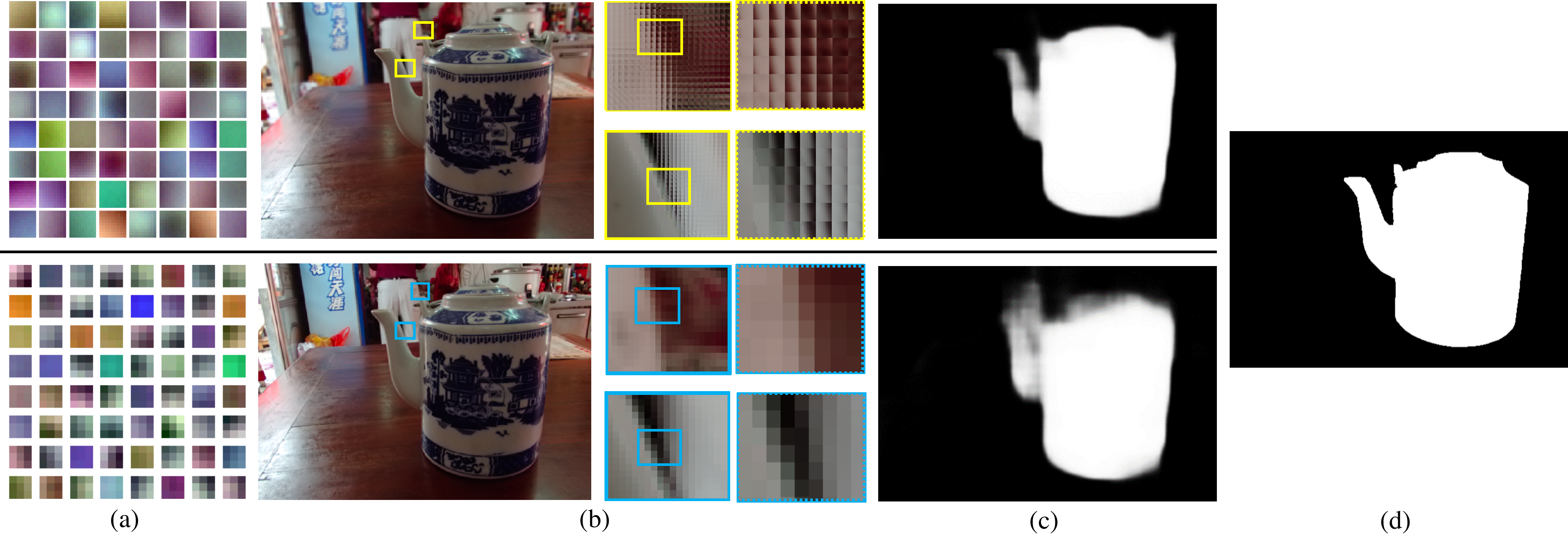}
\caption{Visual comparison of our 4D model (top) and 2D model using the central view (bottom). (a) Visualization of the first {\it conv} layers. (b) Light field input with highlighted regions. (c) Saliency predictions. (d) Ground truth maps.}
\label{fig:kernel}
\end{figure*}
As illustrated in the figure, these variants can separate the most salient regions from similar or cluttered backgrounds.
Compared to other variants, MAC block-$9\times 9$ outputs cleaner and more consistent predictions for the regions with specular reflections (row 1), small salient objects (row 2), and similar foreground and background (rows 2 and 3).
Moreover, we can see that MAC block-$9\times 9$ better predicts salient regions without being highly affected by the light source (row 4).
These results demonstrate that the proposed network variants are likely to extract potential depth cues by learning angular changes, which are helpful to saliency detection.
The kernels with the same size of the angular resolution show better capability in depth discrimination.

\subsection{Model analysis}
Here, we perform all following experiments using MAC block-$9\times 9$, since this setup performed best in previous evaluation.

\subsubsection{Effectiveness of the MAC block}
To further delve into the difference between regular image saliency and light field saliency, we present some important properties of light field features that can better facilitate saliency detection.
We compare our 4D light field saliency (\emph{i.e.} MAC block-$9\times 9$) to 2D model using the central viewing image as input (2D-central view).
The quantitative results are shown in Tables~\ref{comp_2d}.
\begin{table}[!t]
\renewcommand{\arraystretch}{1.3}
\caption{Quantitative comparison between our 4D model and 2D-central view on the proposed Lytro Illum dataset}
\label{comp_2d}
\centering
\begin{tabular}{|l|c|c|c|c|c|}
\hline
Method                                         & F-measure       & WF-measure      & MAE             & AP              \\
\hline
Ours                  & \textbf{0.8116}    & \textbf{0.7540}    & \textbf{0.0551} & \textbf{0.9124}    \\
2D-central view                       & 0.8056 & 0.7446 & 0.0597          & 0.9016 \\
\hline
\end{tabular}
\end{table}
We observe that light field saliency detection with multi-views turns out to perform better than the 2D detector with only the central view.

To provide complementary insight of why light field saliency works, we visualize the weights of the first {\it conv} layers of our network and 2D-central view in Figure~\ref{fig:kernel}(a) to compare angular and spatial patterns.
We can see that the learned weights from our MAC block have noticeable changes in angular space, which suggests that the viewpoint cue of light field data is well captured.
The angular changes are also consistent with the viewpoint variations of micro-lens images, as shown in Figure~\ref{fig:kernel}(b).
The results are attributed to the newly designed {\it conv} method in which the kernel size is the same as the angular resolution of the micro-lens image, and the stride length guarantees angular features are extracted for each micro-lens image.
Therefore, our 4D saliency detector produces more accurate saliency maps than the 2D detector shown in Figure~\ref{fig:kernel}(c).

\begin{figure}[!t]
\centering
\includegraphics[width=1\linewidth]{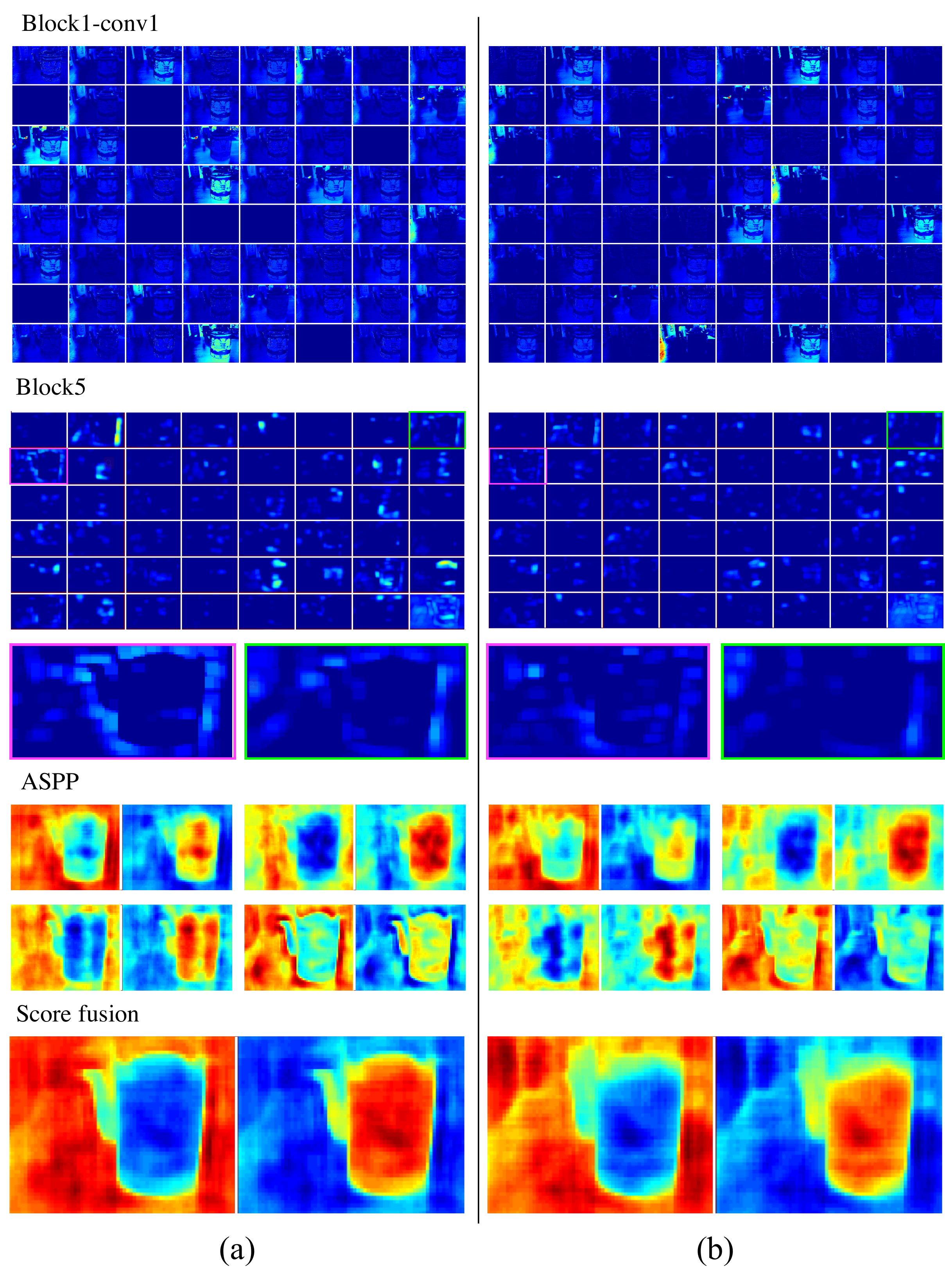}
\caption{Feature maps obtained from (a) 4D model and (b) 2D model using the central view from different layers. From top to bottom: the first {\it conv} features of block1, block5 output features, ASPP features with four atrous rates, and the score fusion maps via sum-pooling. For ASPP and sum fusion, the non-salience and salience scores are shown in the left and right subfigures, respectively.}
\label{fig:features}
\end{figure}

\begin{figure}[!t]
\centering
\includegraphics[width=1\linewidth]{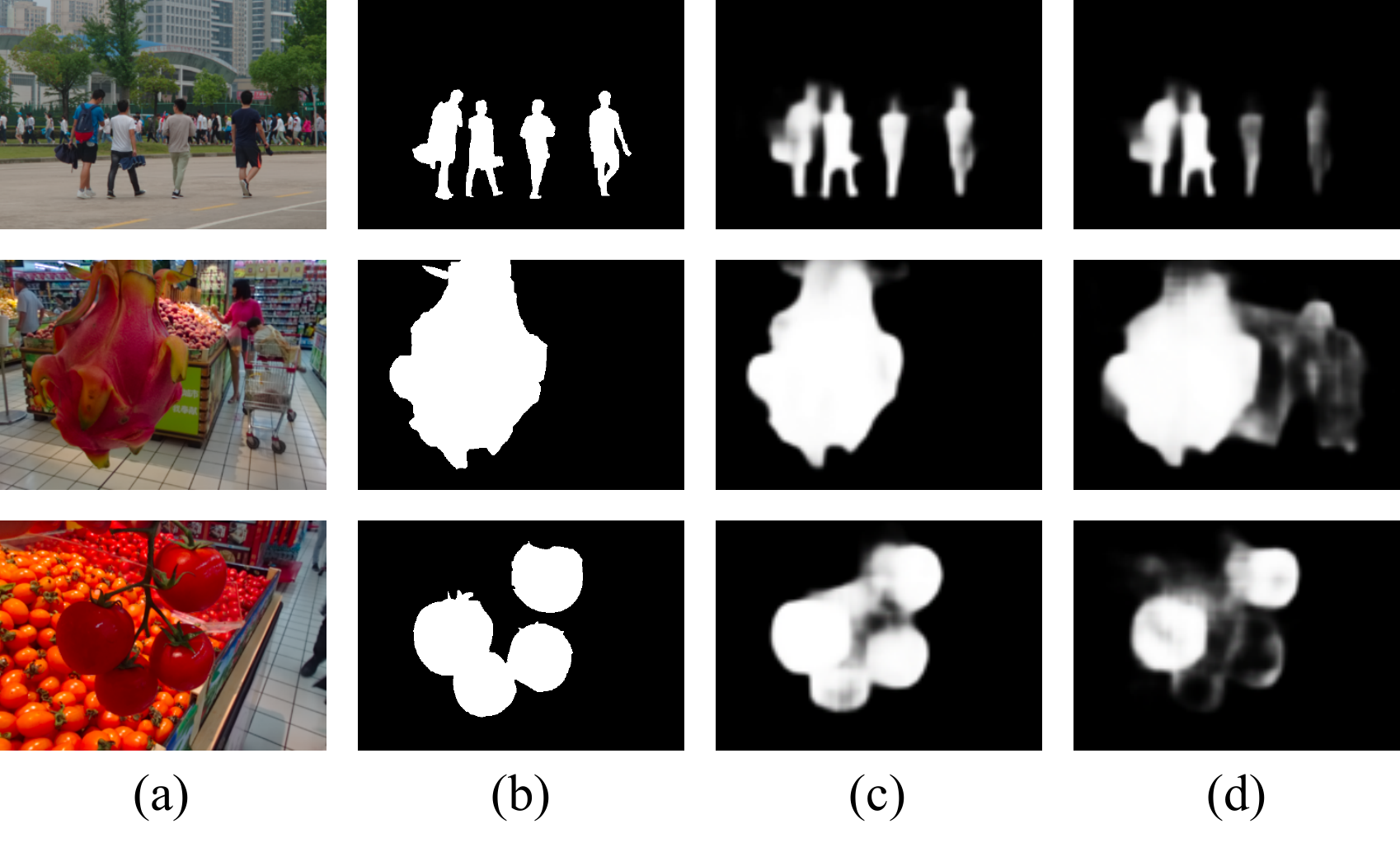}
\caption{Qualitative comparison of 4D model and 2D-central view. (a) Central viewing images. (b) Ground-truth maps. (c) Ours. (d) 2D-central view.}
\label{fig:comp_2d}
\end{figure}

In addition, we show the feature maps obtained from the two models in Figure~\ref{fig:features}.
It can be seen that different layers encode different types of features. 
Higher layers capture semantic concepts of the salient region, whereas lower layers encode more discriminative features for identifying the salient region.
The proposed 4D saliency detector can well discriminate the white spout from the white pants, as shown in Figure~\ref{fig:features}(a).
However, as illustrated in the block1-conv1 and block5 of Figure~\ref{fig:features}(b), most feature maps from the 2D detector have small values that are not discriminative enough to separate the salient tea cup from the pants.
Thus the 2D detector produces features cluttered with background noise in the following ASPP and score fusion.
More comparisons of saliency maps between the two models can be seen in Figure~\ref{fig:comp_2d}.

\subsubsection{Effect of the angular resolutions}
To show the effect of the angular resolutions in the network, we compare the performance of our architecture with varying number of viewpoints in Table~\ref{angular}.
\begin{table}
\renewcommand{\arraystretch}{1.3}
\caption{Effects of the angular resolution on the proposed dataset}
\label{angular}
\centering
\begin{tabular}{|l|c|c|c|c|c|}
\hline
Angular resolution                               & F-measure       & WF-measure      & MAE             & AP              \\
\hline
$7\times 7$                    & 0.8018   & 0.7406    & 0.0567 & \textbf{0.9135}    \\
$9\times 9$                  & \textbf{0.8116}          & \textbf{0.7540}         & \textbf{0.0551}          & 0.9124          \\
$11\times 11$                              & 0.8006          & 0.7392          & 0.0567    & 0.9109         \\
\hline
\end{tabular}
\end{table}
Note that we change the kernel size to stay the same with the angular resolution.
From the table, we can see that the network using $9\times 9$ viewpoints shows the best performance overall.
Increasing the angular resolution to $11 \times 11$ cannot improve the performance, which can be explained by the fact that the viewing angles at the boundary are very oblique~\cite{Levoy09} and the narrow baseline of the light field camera leads to high viewing redundancy with higher angular resolutions~\cite{Levoy96,Pendu18}.

\subsubsection{Overfitting issues}
\begin{figure*}[!t]
\centering
\includegraphics[width=1\linewidth]{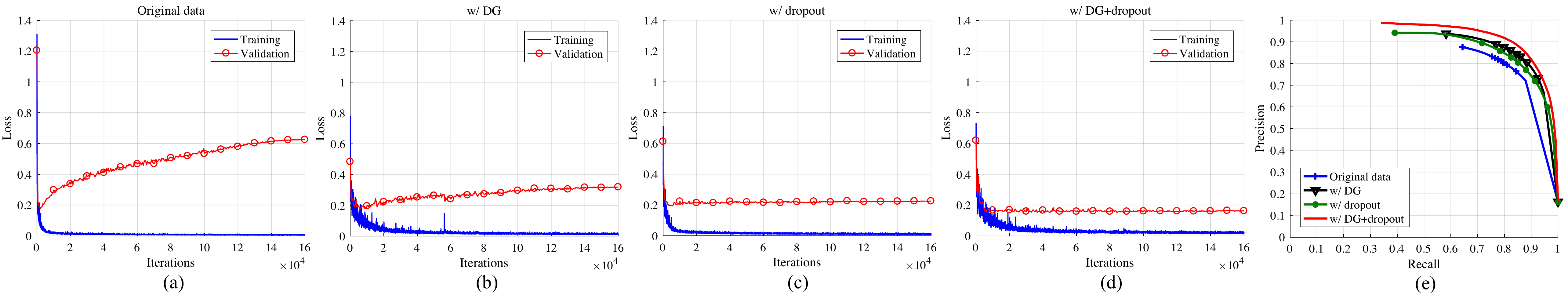}
\caption{Training and validation loss for our model on the proposed Lytro Illum dataset. (a) Original training data. (b) Training with DG. (c) Training with dropout. (d) Training with DG and dropout. (e) PR curves for different strategies.}
\label{fig:overfit}
\end{figure*}

Overfitting is a common problem related to training a CNN with limited data.
In this section, we analyse the proposed network by introducing different strategies to handle overfitting: data augmentation (DG) and dropout. 
The results obtained for our best performing model are shown in Figure~\ref{fig:overfit}.
Clearly, the network is overfitting with original training data as shown in Figure~\ref{fig:overfit}(a).
As expected, both DG and dropout are crucial to minimize overfitting as shown in Figures~\ref{fig:overfit}(b)--(d).
Figure~\ref{fig:overfit}(e) presents the corresponding PR curves.
It can be seen that by increasing the amount and diversity of the data and the amount of dropout between different layers during training, the performance of the network increases as well.

\begin{table*}
\renewcommand{\arraystretch}{1.3}
\caption{Quantitative comparison of our approach and other 2D models on the proposed dataset. Bold: best, underlined: second best}
\label{2D}
\centering
\begin{tabular}{|l|c|c|c|c|c|}
\hline
					& Model                               & F-measure       & WF-measure      & MAE             & AP              \\
\hline
\multirow{2}{*}{Traditional}	 &MST~\cite{Tu16}                    & 0.6695   & 0.5834    & 0.1243 & 0.6967    \\
					 &SMD~\cite{Peng17}                  & 0.7246          & 0.5371        & 0.1234          & 0.7789          \\
					 &MDC~\cite{Huang17}               & 0.7407          & 0.5891          & 0.1094    & 0.7552        \\
					 &WFD~\cite{Huang18}		& 0.7260          & 0.6408       & 0.1024    & 0.7604    \\
\hline
\multirow{2}{*}{CNN-based} &PiCANet~\cite{Liu18}      & 0.7908              & 0.6745           & 0.0782      & 0.8362     \\
					   &Amulet~\cite{Zhang17b}     & {\ul 0.8059}            & \textbf{0.7686}           & {\ul 0.0552}      & {\ul 0.8485}     \\
					 &LFR~\cite{Zhang18a}         & 0.7756             & 0.7242          & 0.0702       & 0.8463    \\
					   &HyperFusion~\cite{Zhang19b}    & 0.7549     &0.6945          & 0.0752       &0.8359    \\
					   & Ours                            & \textbf{0.8116}    & {\ul 0.7540}    & \textbf{0.0551} & \textbf{0.9124}    \\
\hline 
\end{tabular}
\end{table*}
\subsection{Comparison with 2D models}
To understand the additional information contained in the micro-lens light field images, we compare our best performing approach (\emph{i.e.} MAC block-$9\times 9$) to 8 existing methods on the test set of our proposed dataset.
Our comparison includes 4 traditional approaches MST~\cite{Tu16}, SMD~\cite{Peng17}, MDC~\cite{Huang17}, WFD~\cite{Huang18}; and 4 CNN-based ones: PiCANet~\cite{Liu18}, Amulet~\cite{Zhang17b}, LFR~\cite{Zhang18a}, HyperFusion~\cite{Zhang19b}. 
To facilitate fair comparison and effective model training, we use the recommended parameter settings provided by the authors to initialize these models.
All CNN-based methods are based on DNNs pre-trained on the ImageNet~\cite{Russakovsky15} classification task.
We retrain these CNN models on the proposed dataset in a five-fold cross-validation way and apply the same data augmentation method used in our work.
The quantitative results are shown in Table~\ref{2D}.
We can see that in general, our model outperforms other methods in terms of F-measure, MAE, and AP metrics.
Amulet~\cite{Zhang17b} obtains the second best performance on the proposed dataset.
CNN-based methods consistently perform better than traditional methods.

\subsection{Comparison to state-of-the-art light field methods}
\begin{figure*}[!t]
\centering
\includegraphics[width=1\linewidth]{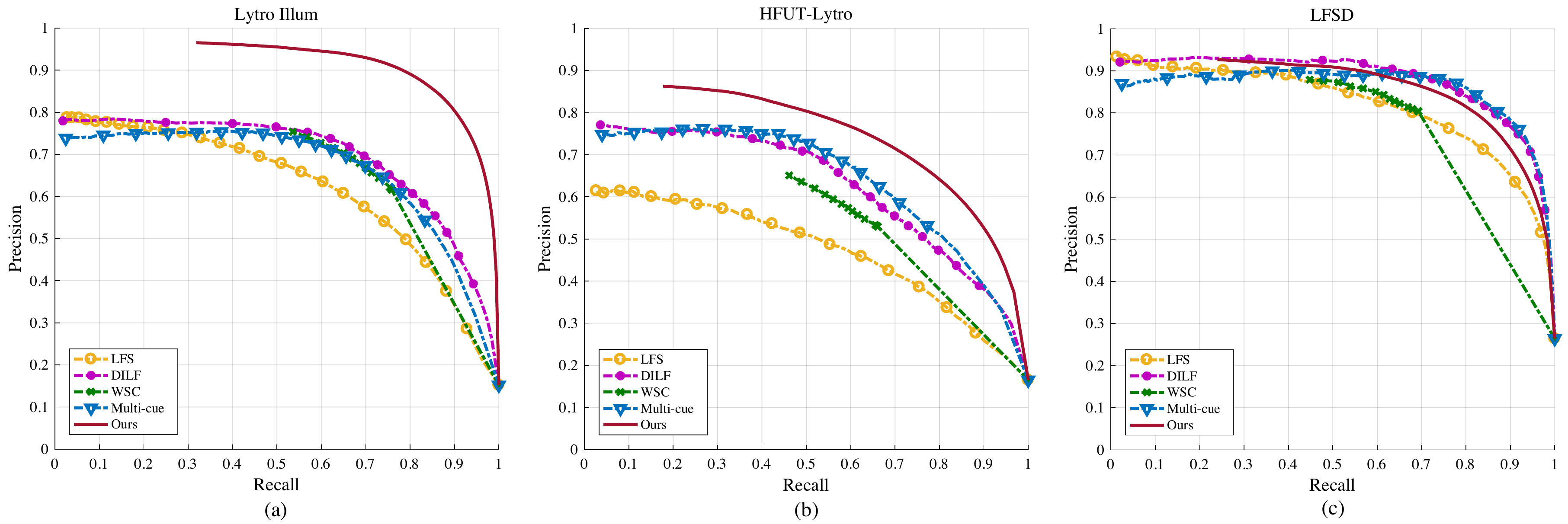}
\caption{Comparison on three datasets in terms of PR curve. (a) The proposed Lytro Illum dataset. (b) The HFUT-Lytro dataset. (c) The LFSD dataset.}
\label{fig:pr}
\end{figure*}

We compare our best performing model MAC block-$9\times 9$ to four state-of-the-art methods tailored to light field saliency detection: Multi-cue~\cite{Zhang17}, DILF~\cite{Zhang15}, WSC~\cite{Li15}, and LFS~\cite{Li17}.
We train our network on the novel dataset, and evaluate on the others without fine-tuning.
The results of other methods are obtained using the authors' implementations.
Tables~\ref{illum}--\ref{lfsd} and Figure~\ref{fig:pr} show quantitative results on three datasets.
Overall, our approach outperforms other methods on three datasets without any post-processing for refinement, which demonstrates the advantage of the proposed deep convolutional network for light field saliency detection.
In particular, we observe that the proposed approach shows significant performance gains when compared to previous methods on the proposed dataset for all metrics.
The performance is lower on the HFUT-Lytro and LFSD datasets, which is due to the limited viewpoint information in these datasets.
Therefore, a large number of filters learnt on the proposed dataset are underused.
This demonstrates that different light field datasets do affect the accuracy of methods.
Multi-cue~\cite{Zhang17} and DILF~\cite{Zhang15} methods show better performance than our approach in terms of F-measure and AP on the LFSD dataset.
The reason is that these methods use external depth features and post-processing refinement to improve the performance.
\begin{table}[!t]
\renewcommand{\arraystretch}{1.3}
\caption{Quantitative results on the proposed Lytro Illum dataset. Bold: best, underlined: second best}
\label{illum}
\centering
\begin{tabular}{|l|c|c|c|c|c|}
\hline
Method                                         & F-measure       & WF-measure      & MAE             & AP              \\
\hline
LFS~\cite{Li17}          & 0.6107          & 0.3596          & 0.1697          & 0.6193         \\
WSC~\cite{Li15}          & 0.6451          & \uline{0.5945}          & \uline{0.1093}          & 0.5958          \\
DILF~\cite{Zhang15}      & 0.6395          & 0.4844          & 0.1389          & \uline{0.6921}          \\
Multi-cue~\cite{Zhang17} & \uline{0.6648}          & 0.5420          & 0.1197          & 0.6593          \\
\hline
Ours                  & \textbf{0.8116}    & \textbf{0.7540}    & \textbf{0.0551} & \textbf{0.9124}    \\
\hline
\end{tabular}
\end{table}
\begin{table}[!t]
\renewcommand{\arraystretch}{1.3}
\caption{Quantitative results on the HFUT-Lytro dataset. Bold: best, underlined: second best}
\label{lytro}
\centering
\begin{tabular}{|l|c|c|c|c|c|}
\hline
Method                                         & F-measure       & WF-measure      & MAE             & AP              \\
\hline
LFS~\cite{Li17}          & 0.4868          & 0.3023          & 0.2215          & 0.4718         \\
WSC~\cite{Li15}          & 0.5552          & 0.5080          & 0.1454          & 0.4743          \\
DILF~\cite{Zhang15}      & 0.5543          & 0.4468         & 0.1579          & 0.6221         \\
Multi-cue~\cite{Zhang17} & \uline{0.6135}          & \uline{0.5146}         & \uline{0.1388}          & \uline{0.6354}          \\
\hline
Ours                 & \textbf{0.6721}    & \textbf{0.6087}   & \textbf{0.1029} & \textbf{0.7390}    \\
\hline
\end{tabular}
\end{table}
\begin{table}[!t]
\renewcommand{\arraystretch}{1.3}
\caption{Quantitative results on the LFSD dataset. Bold: best, underlined: second best}
\label{lfsd}
\centering
\begin{tabular}{|l|c|c|c|c|c|}
\hline
Method                                         & F-measure       & WF-measure      & MAE             & AP              \\
\hline
LFS~\cite{Li17}          & 0.7525          & 0.5319         & 0.2072          & 0.8161         \\
WSC~\cite{Li15}          & 0.7729          & \uline{0.7371}          & 0.1453          & 0.6832          \\
DILF~\cite{Zhang15}      & \uline{0.8173}          & 0.6695        & \uline{0.1363}         & \textbf{0.8787}         \\
Multi-cue~\cite{Zhang17} & \textbf{0.8249}          & 0.7155         & 0.1503          & \uline{0.8625}         \\
\hline
Ours                    & 0.8105    & \textbf{0.7378}    & \textbf{0.1164} & 0.8561    \\
\hline
\end{tabular}
\end{table}

\begin{figure*}[!t]
\centering
\includegraphics[width=1\linewidth]{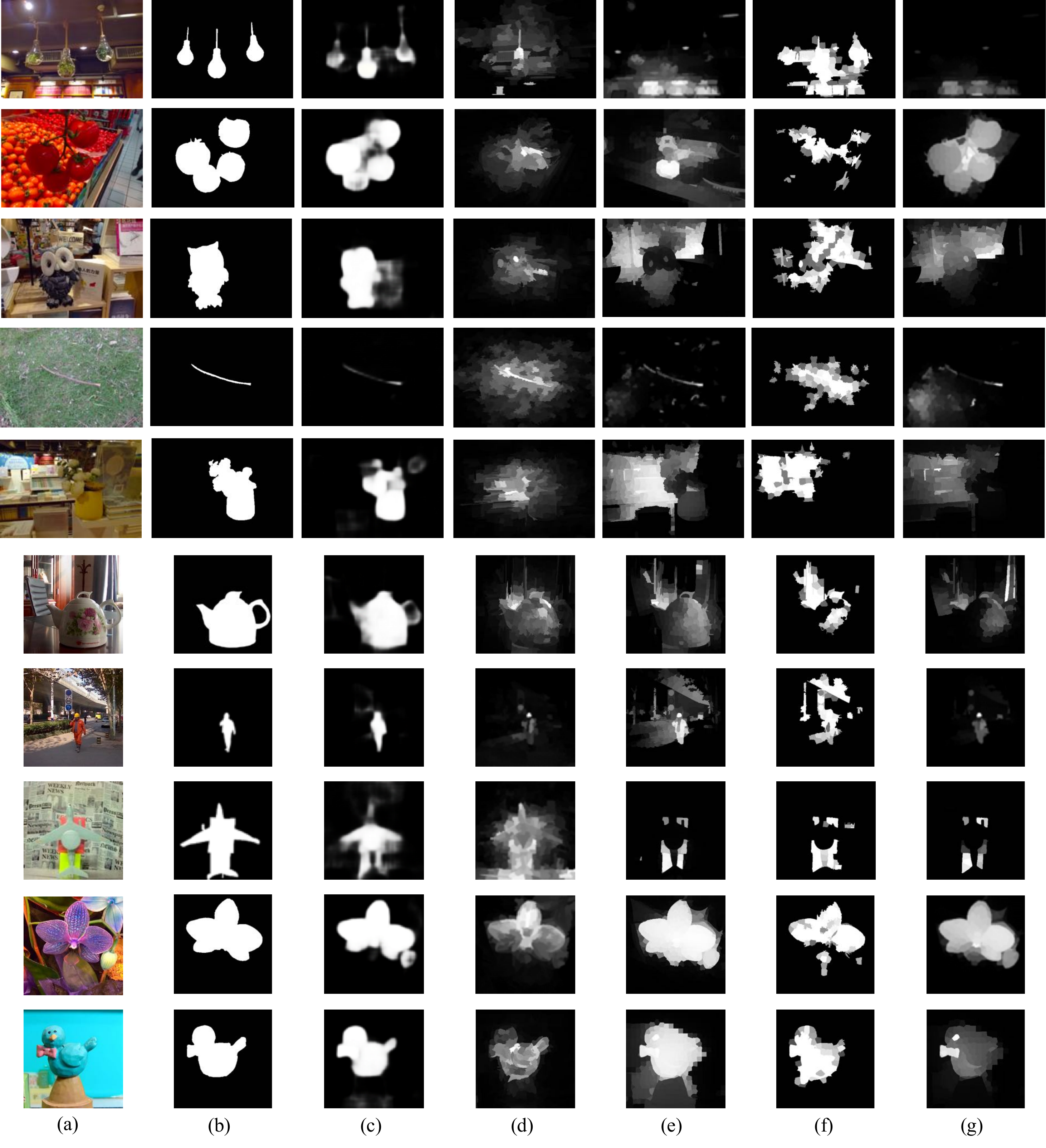}
\caption{Visual comparison of our best MAC block variant (Ours) and state-of-the-art methods on three datasets. (a) Central viewing/all-focus images. (b) Ground truth maps. (c) Ours. (d) LFS~\cite{Li14}. (e) DILF~\cite{Zhang15}. (f) WSC~\cite{Li15}. (g) Multi-cue~\cite{Zhang17}. The first five samples are taken from the proposed Lytro Illum dataset, the middle three samples are taken from the HFUT-Lytro dataset, and the last two samples are taken from the LFSD dataset.}
\label{fig:vis}
\end{figure*}

Some qualitative results are shown in Figure~\ref{fig:vis}.
We can see that our approach can handle various challenging scenarios, including multiple salient objects (rows 1 and 2), cluttered backgrounds (rows 3 and 5), small salient objects (rows 4 and 7), inconsistent illumination (rows 1 and 6), and salient objects in similar backgrounds (rows 8, 9 and 10).
It is also worth noting that without any post-processing, our approach can highlight salient objects more uniformly than other methods.

\section{Conclusion} \label{sec:conclusion}
This paper introduces a deep convolutional network for saliency detection on light fields by exploiting multi-view information in micro-lens images.
Specifically, we propose MAC block variants to process the micro-lens image array.
This paper can be viewed as the first work that addresses light field saliency detection using an end-to-end CNN.
To facilitate training such a deep network, we introduce a challenging saliency dataset with light field images captured from a Lytro Illum camera.
In total, 640 high quality light fields are produced, making the dataset the largest among existing light field saliency datasets. 
Extensive experiments demonstrate that comparing to 2D saliency based on the central view alone, 4D light field saliency can exploit additional angular information contributing to an increase in the performance of saliency detection.
The proposed network is superior to saliency detection methods designed for 2D RGB images on the proposed dataset, and outperforms the state-of-the-art light field saliency detection methods on the proposed dataset and generalizes well to the existing datasets.
In particular, our approach is capable of detecting salient regions in challenging cases, such as with similar foregrounds and backgrounds, inconsistent illumination, multiple salient objects, and cluttered backgrounds.
Our work suggests promising future directions of exploiting spatial and angular patterns in light fields and deep learning technologies to advance the state-of-the-art in pixel-wise prediction tasks.

\ifCLASSOPTIONcaptionsoff
  \newpage
\fi



%

\bibliographystyle{IEEEtran}
\bibliography{egbib}

%
%

%




\end{document}